\documentclass[letterpaper, 10 pt, conference]{ieeeconf}  % Comment this line out if you need a4paper

\IEEEoverridecommandlockouts                              % This command is only needed if 
\usepackage{graphics} % for pdf, bitmapped graphics files
\usepackage{epsfig} % for postscript graphics files
\usepackage{times} % assumes new font selection scheme installed
\usepackage{amsmath} % assumes amsmath package installed
\usepackage{amssymb}  % assumes amsmath package installed
\usepackage{multirow}

\usepackage{booktabs}
\usepackage{url}
\usepackage{cite}
\usepackage{xcolor}
\usepackage[
    colorlinks=true, % 必须设置为 true，否则文字是黑色，只有边框有颜色
    linkcolor=red,   % 内部链接颜色（如目录、\ref 交叉引用）
    urlcolor=blue,   % 外部网址颜色（\href 和 \url 使用此颜色）
    citecolor=green  % 参考文献引用颜色（\cite 使用此颜色）
]{hyperref}

\title{\LARGE \bf
LAG-IO: Multi-IMU State Estimation for Legged Robots \\ Using Factor Graphs
}

\author{Yicheng Zhou  \and Longwen Chen \and Qinghua Yu \and Junhao Xiao \and Huimin Lu\and  Xieyuanli Chen % <-this % stops a space
}

\begin{document}

\setlength{\abovedisplayskip}{3pt}
\setlength{\belowdisplayskip}{3pt}
\setlength{\abovedisplayshortskip}{0pt}
\setlength{\belowdisplayshortskip}{0pt}

\maketitle
\thispagestyle{empty}
\pagestyle{empty}

%%%%%%%%%%%%%%%%%%%%%%%%%%%%%%%%%%%%%%%%%%%%%%%%%%%%%%%%%%%%%%%%%%%%%%%%%%%%%%%%
\begin{abstract}
Robust proprioceptive state estimation is crucial for legged robots, particularly in challenging environments where exteroceptive sensors are unreliable. While traditional leg odometry typically relies on a single body-mounted inertial measurement unit (IMU) fused with leg kinematics via an extended Kalman filter (EKF), such methods are vulnerable to linearization errors and hard-contact assumptions, leading to severe drift during slippage or dynamic locomotion.
%%%%%
In this paper, we propose LAG-IO, a factor graph-based multi-IMU state estimation framework that tightly couples measurements from a main body IMU, multiple leg-mounted IMUs, and joint encoders within a sliding-window optimization. To handle non-stationary foot-ground interactions, we introduce a continuous probabilistic contact detection module based on kernel density estimation (KDE), which adaptively scales the covariance of zero-velocity update (ZUPT) factors. Furthermore, kinematic relative position factors are formulated to structurally anchor the independent leg inertial states to the robot's main body. Experimental results on a quadruped robot across various unstructured terrains demonstrate that, compared to existing baselines, LAG-IO achieves higher planar trajectory accuracy, reduces unobservable vertical divergence, and provides smoother trajectory transitions during unstable contacts, while maintaining competitive attitude observability. The code is released at: \href{https://github.com/nubot-nudt/LAG-IO}{https://github.com/nubot-nudt/LAG-IO}.
\end{abstract}

%%%%%%%%%%%%%%%%%%%%%%%%%%%%%%%%%%%%%%%%%%%%%%%%%%%%%%%%%%%%%%%%%%%%%%%%%%%%%%%%
\section{Introduction}

Accurate proprioceptive state estimation is crucial for legged robots to execute dynamic locomotion and navigation tasks. While exteroceptive sensors such as LiDAR and cameras provide rich environmental information for localization, their performance can degrade significantly in adverse weather conditions or visually impoverished environments~\cite{zhao2025resilient, kuang2025reslpr}. Thus, a robust proprioceptive estimator is required to provide reliable, high‑frequency state feedback when external perception becomes unreliable or fails.

Existing proprioceptive state estimation methods mainly rely on fusing measurements from a single body-mounted IMU and joint kinematics~\cite{bloesch2013state, hartley2020contact, wisth2019robust}. Traditional filtering frameworks, such as the extended Kalman filter, typically employ zero-velocity updates (ZUPT) under the assumption that the stance foot maintains a rigid, stationary contact with the ground \cite{wahlstrom2020fifteen}. However, when foot slips occur on unstructured terrains, this no-slip assumption is violated, injecting erroneous biases that inevitably lead to trajectory drift \cite{kim2022step}. To mitigate these slip-induced errors, recent literature has explored end-to-end learning-based odometry networks \cite{liu2020tlio, buchanan2022learning} and neural-augmented filters \cite{gu2024enhancing, lee2025legged} to compensate for complex, unmodeled contact dynamics. While these data-driven approaches can capture intricate motion patterns, they often struggle with generalizability across unseen environments and remain strictly bounded by their training datasets.

Expanding the proprioceptive sensing suite from a single IMU to a multi‑IMU configuration has recently emerged as an effective way to address the limitations of traditional setups. Placing additional IMUs on the lower limbs allows the system to capture local impact dynamics directly, avoiding dependence on fragile force sensors~\cite{yang2023multi}. Building on this idea, the recent DogLegs framework~\cite{wu2025doglegs} showed that combining a body IMU, multiple leg IMUs, and kinematic constraints within an EKF can markedly improve leg odometry performance. However, despite these hardware gains, the EKF remains limited by its recursive structure. Once a measurement is incorporated, the associated linearization point cannot be revisited. This constraint prevents the filter from correcting past linearization errors, making it vulnerable to transient nonlinear disturbances caused by unstable contacts and high‑frequency impacts.

To address these limitations, we propose LAG-IO, which reformulates multi-IMU kinematic fusion within a tightly-coupled, sliding-window factor graph optimization. Unlike filtering methods~\cite{bloesch2013state, hartley2020contact, lee2025legged}, the graph-based backend maintains a recent history of states, naturally enabling iterative relinearization to smooth out slip-induced outliers. Within this unified graph, continuous inertial preintegration and kinematic relative position (RelPos) factors are incorporated to structurally anchor the independent leg states to the robot's main body. Furthermore, a continuous contact probability evaluated from leg-IMU signals is proposed to dynamically scale the variance of the ZUPT factors, gracefully relaxing the zero-velocity constraints during unstable contacts.
% The main contributions of this work are summarized as follows:
% \begin{itemize}
%     \item We propose LAG-IO, a tightly-coupled sliding-window factor graph optimization framework for multi-IMU legged robots, which provides enhanced resilience against nonlinear drift and slip-induced outliers during dynamic locomotion.
%     \item We extensively evaluate the proposed framework on a public multi-IMU legged locomotion dataset. Experimental results demonstrate that our graph-based approach achieves highly competitive performance and enhanced robustness compared to existing multi-IMU filtering baseline.
% \end{itemize}
The main contributions of this work are threefold:
\begin{itemize}
    \item We propose LAG-IO, a novel sliding-window factor graph optimization framework for multi‑IMU legged proprioception, integrating body and leg inertial states with kinematics‑based relative position constraints to ensure consistent cross‑limb anchoring and improved robustness over recursive filtering.
    \item We propose a continuous probabilistic contact weighting mechanism that computes a stationarity likelihood from leg‑IMU motion statistics and adaptively scales ZUPT covariances, enabling graceful handling of micro‑slips, soft contacts, and impact disturbances.
    \item We provide a real‑time, impact‑resilient implementation with noise‑shaped leg‑IMU preintegration and robust optimization, achieving higher planar accuracy, reduced vertical drift, and smoother trajectory transitions compared to the EKF baselines. The code will be open source to benefit future research.
\end{itemize}

%%%%%%%%%%%%%%%%%%%%%%%%%%%%%%%%%%%%%%%%%%%%%%%%%%%%%%%%%%%%%%%%%%%%%%%%%%%%%%%%
\section{Related Works}

Proprioceptive state estimation for legged robots relies on the continuous dead-reckoning of inertial measurements bounded by intermittent kinematic constraints. The evolution of this field has been driven by the dual requirements of resolving mathematical inconsistencies in state propagation and managing the physical uncertainties inherent in dynamic foot-ground interactions.

The foundational approach to legged proprioception, established by Bloesch et al. \cite{bloesch2013state}, formulated an EKF to jointly estimate the main body state and foot contact positions. This architecture applies state updates using forward kinematics as pseudo-measurements whenever a stable contact is detected. However, standard EKFs operating on the $SE(3)$ manifold suffer from spurious observability, where state-dependent Jacobians can inadvertently inject erroneous information into unobservable directions such as global yaw. Researchers subsequently introduced Lie group symmetries into the filtering process. The resulting Invariant EKF (InEKF) \cite{hartley2020contact, hohmeyer2025inekformer} ensures that the error dynamics of the filter remain independent of the current state trajectory. Although the InEKF resolves these structural symmetry issues, its estimation accuracy remains sensitive to boolean contact assumptions. Unmodeled foot slippage in unstructured environments violates the rigid kinematic measurement model and may subsequently induce progressive trajectory drift.

Data-driven methodologies offer an alternative by reducing the reliance on analytical kinematic models during uncertain contacts. Early applications primarily utilized machine learning as heuristic classifiers to replace threshold-based contact detection \cite{camurri2017probabilistic}. Recent end-to-end architectures, such as TLIO \cite{liu2020tlio}, leverage temporal convolutional networks to directly regress 3D displacements from local IMU sequences, thereby bypassing the kinematic chain entirely. Concurrently, hybrid approaches \cite{gu2024enhancing, lee2025legged} employ recurrent neural networks to dynamically estimate pseudo-measurement noise covariances or contact biases. This allows the filter to adaptively de-weight unreliable kinematic updates. While learning-augmented methods can model complex impact dynamics, their zero-shot generalizability is often constrained when the robot encounters out-of-distribution terrains or novel gait patterns not present in the training datasets.

A fundamental structural limitation of recursive filtering is that the linearization point is permanently fixed upon each measurement update. The community has increasingly investigated smoothing and optimization-based backends to address this constraint. Fixed-lag smoothers \cite{kim2022step} maintain a short history of states, enabling the system to retroactively re-evaluate the trajectory window upon detecting slip events. Expanding upon this concept, FGO reformulates the estimation as a Maximum A Posteriori inference problem. As demonstrated by Wisth et al. \cite{wisth2019robust}, maintaining a sliding window of historical states allows for iterative relinearization. This optimization process provides a robust mechanism to reject delayed outlier measurements and accommodate non-linear impact shocks.

Parallel to algorithmic advancements, enriching the physical observation model through multi-sensor fusion has emerged as a practical frontier. Due to the compliance and damping characteristics of legged structures, high-frequency foot-ground interaction forces are significantly attenuated before reaching a single body-mounted IMU. Deploying additional IMUs on the lower limbs allows for the direct measurement of these localized dynamics \cite{yang2023multi}. The DogLegs framework \cite{wu2025doglegs} demonstrated that tightly fusing a primary body IMU with multiple leg IMUs enhances the observability of the kinematic chain. However, because DogLegs is architecturally built upon an EKF, it can still struggle to fully recover from transient linearization errors during severe and unstable contacts. Building upon this multi-IMU paradigm, the proposed LAG-IO framework migrates the fusion engine to a sliding-window factor graph. 
By incorporating continuous KDE-based contact probabilities to adaptively scale the ZUPT covariances, LAG-IO preserves the physical observability of distributed inertial sensing while exploiting the resilience of iterative optimization.

%%%%%%%%%%%%%%%%%%%%%%%%%%%%%%%%%%%%%%%%%%%%%%%%%%%%%%%%%%%%%%%%%%%%%%%%%%%%%%%%
\section{Our Approach}

\begin{figure*}[t]
    \centering
    \includegraphics[width=0.9\textwidth]{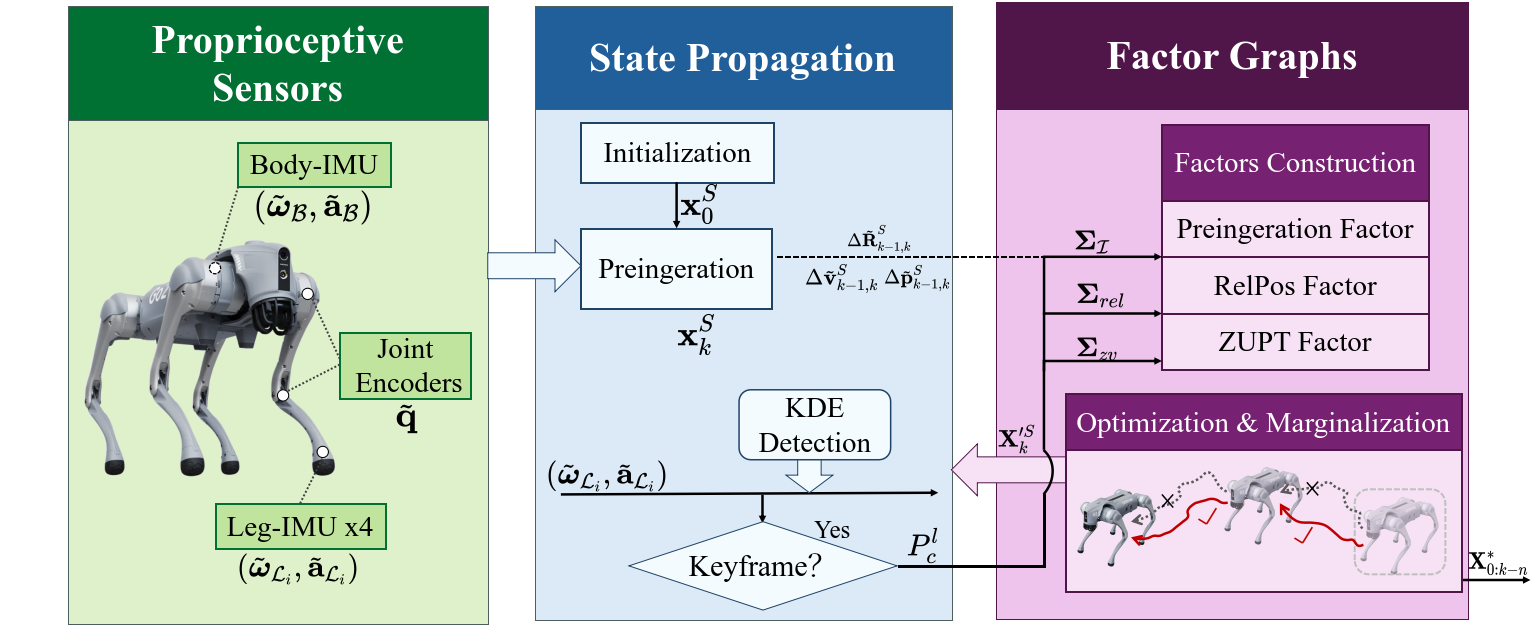}
    \vspace{-0.3cm}
    \caption{System architecture of the LAG-IO framework. Proprioceptive measurements from the body-IMU, four leg-IMUs, and joint encoders (hip, thigh, and calf) are processed through state preintegration and KDE-based contact detection to construct the tightly-coupled factor graph. The multi-IMU states are subsequently solved via sliding-window optimization and bounded by marginalization. The grey dashed box in the bottom-right indicates the oldest keyframe undergoing marginalization, while the grey dotted and red solid trajectories denote the drifting prior and the optimized estimates.}
    \label{fig:system_architecture}
    \vspace{-0.5cm}
\end{figure*}

The proposed LAG-IO framework processes multiple sensor streams within a unified sliding-window graph optimization. The overall system architecture is illustrated in Fig.~\ref{fig:system_architecture}. Rather than recursively compressing past measurements, the graph-based backend dynamically instantiates keyframes driven by a probabilistic contact detection mechanism. These temporally distributed keyframes are subsequently constrained by inertial preintegration, probability-scaled ZUPT factors, and kinematic RelPos factors.

\subsection{Definition of Coordinate System and State}

\begin{figure*}[t]
    \centering
    \includegraphics[width=0.8\textwidth]{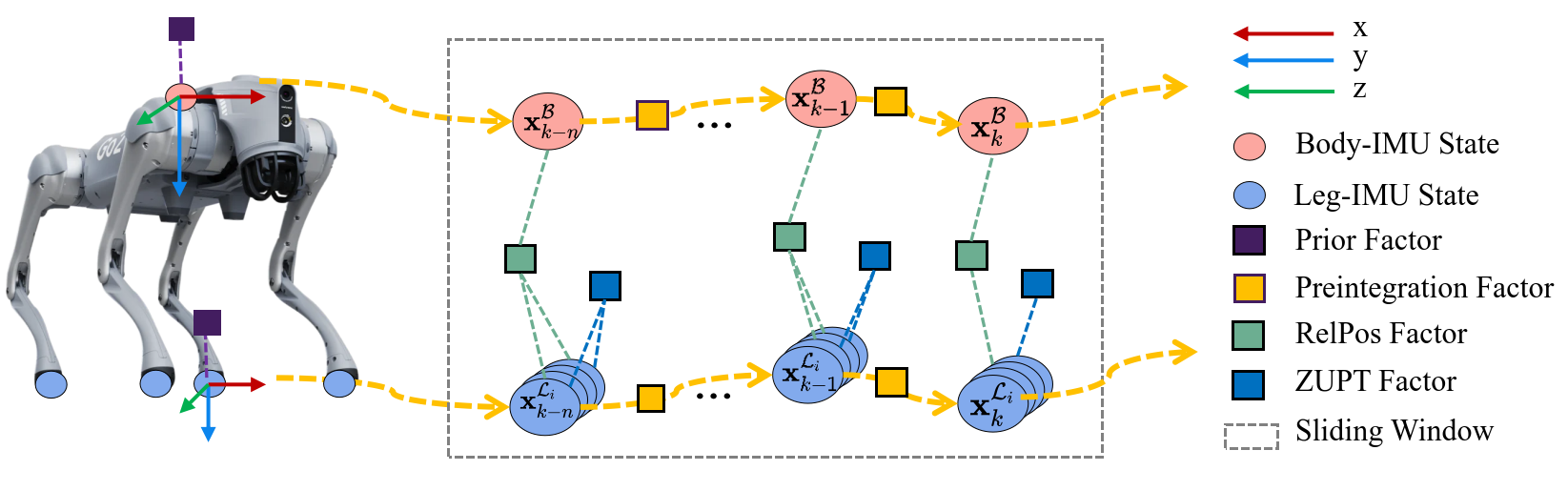} 
    \vspace{-0.5cm}
    \caption{Topology of the LAG-IO factor graph and sensor frames. Local moving frames follow the Front-Right-Down (FRD) convention. IMU preintegration factors (yellow squares) continuously constrain all body (pink circles) and leg (blue circles) states within the sliding window. Conversely, at each keyframe, relative position (green squares) and zero-velocity (blue squares) factors are applied exclusively to the specific detected contact legs. For instance, at keyframe $k$, these constraints are  added to only a single active stance leg.}
    \label{fig:factor_topology}
    \vspace{-0.5cm}
\end{figure*}

Rigorous spatial referencing is a prerequisite for multi-sensor state estimation. Let $\mathcal{W}$ denote the global inertial navigation frame, which strictly adheres to the North-East-Down (NED) convention, such that the gravity vector $\mathbf{g}$ aligns with the positive z-axis. The local moving frames, as explicitly marked in the upper schematic of Fig. \ref{fig:factor_topology}, consist of the main body frame $\mathcal{B}$ following the Front-Right-Down (FRD) convention, alongside four independent leg-IMU frames $\mathcal{L}_i$, where $i \in \{1, 2, 3, 4\}$.

Within the sliding-window formulation, the full system state $\mathbf{X}_k$ at keyframe $k$ encapsulates the coupled navigation states of both the main body and the lower limbs of the legged robot:
\begin{equation}
    \mathbf{X}_k = \left[ \mathbf{x}_k^{\mathcal{B}}, \mathbf{x}_k^{\mathcal{L}_1}, \mathbf{x}_k^{\mathcal{L}_2}, \mathbf{x}_k^{\mathcal{L}_3}, \mathbf{x}_k^{\mathcal{L}_4} \right]
\end{equation}
For each individual sensor node $S \in \{\mathcal{B}, \mathcal{L}_1, \dots, \mathcal{L}_4\}$, the internal navigation state $\mathbf{x}_k^S$ is parameterized as:
\begin{equation}
    \mathbf{x}_k^S = \left[ \mathbf{p}_k^S, \mathbf{v}_k^S, \mathbf{R}_k^S, \mathbf{b}_{a,k}^S, \mathbf{b}_{g,k}^S \right]
\end{equation}
where $\mathbf{p}_k^S, \mathbf{v}_k^S \in \mathbb{R}^3$ represent the translation and velocity vectors in $\mathcal{W}$, respectively. The orientation matrix $\mathbf{R}_k^S \in \mathrm{SO}(3)$ transforms vectors from the local frame $S$ to the global frame $\mathcal{W}$. The terms $\mathbf{b}_{a,k}^S, \mathbf{b}_{g,k}^S \in \mathbb{R}^3$ denote the accelerometer and gyroscope biases, which are modeled as random walks.

System initialization is executed under stationary conditions before the robot starts moving. During this initial static phase, the initial roll and pitch attitudes and the IMU gyroscope biases are calibrated. The initial relative spatial offsets between the body $\mathcal{B}$ and each leg $\mathcal{L}_i$ are established by averaging the forward kinematics derived from the joint encoder measurements, providing a kinematically consistent initial configuration for the factor graph.

%%%%%%%%%%%%%%%%%%%%%%%%%%%%%%%%%%%%%%%%%%%%%%%%%%%%%%%%%%%%%%%%%%%%%%%%%%%%%%%%
\subsection{KDE-Based Contact Detection}

Reliable foot contact detection is fundamental to the stability of proprioceptive odometry for legged robots. Modeling foot-ground interaction strictly as a binary event is inherently insufficient for unstructured terrains, as contact instability involves continuous dynamic transitions rather than discrete states. In contrast to rigid boolean thresholds and learning-based classifiers with limited terrain generalizability, we utilize the kernel density estimation (KDE) framework \cite{maravgakis2023probabilistic}. This approach quantifies the degree of contact stability through a continuous probabilistic metric, reformulating contact detection as a stationarity test of Leg-IMU signals.

Let $\tilde{\mathbf{a}}_k$ and $\tilde{\boldsymbol{\omega}}_k$ denote the raw linear acceleration and angular velocity measurements from a specific Leg-IMU (denoted as frame $\mathcal{L}$). Motion-induced variations are isolated from the gravity vector via compensation: $\hat{\mathbf{a}}_k = \tilde{\mathbf{a}}_k - (\mathbf{R}_k^{\mathcal{L}})^T \mathbf{g}$. Defining the kinematic measurement vector as $\mathbf{u}_k = [\hat{\mathbf{a}}_k^T, \tilde{\boldsymbol{\omega}}_k^T]^T \in \mathbb{R}^6$, we employ a Gaussian kernel to model the underlying measurement distribution. Under this formulation, the KDE evaluates the probability $P_j$ that the deviation $\delta u_{k,j}$ of axis $j$ falls within a predefined static bandwidth $d_j$:

\begin{equation}
    P_{j} = \Phi\left(\frac{d_{j} - \delta u_{k,j}}{\sigma_{j}}\right) - \Phi\left(\frac{-d_{j} - \delta u_{k,j}}{\sigma_{j}}\right)
\end{equation}
$\Phi(\cdot)$ is the standard normal CDF and $\sigma_j$ is the sensor noise. 

The joint contact probability for the $l$-th leg, $P_c^l$, is aggregated as $P_c^l = \prod_{j=1}^{6} P_{j}^l$. A new keyframe is instantiated at time $k$ according to the event $\mathcal{K}$:
\begin{equation}
    \mathcal{K} = \left\{ 
    \begin{array}{ll}
        \text{True}, & \exists l \in \{1,\dots,4\}: P_c^l > \tau_{stable} \\
        \text{False}, & \text{otherwise}
    \end{array} \right.
\end{equation}
where $\mathcal{K}$ triggers the structural update of the factor graph and the concurrent insertion of kinematic-inertial constraints.
%%%%%%%%%%%%%%%%%%%%%%%%%%%%%%%%%%%%%%%%%%%%%%%%%%%%%%%%%%%%%%%%%%%%%%%%%%%%%%%%
\subsection{IMU Preintegration Factor}

Consecutive keyframes $k$ and $k+1$, triggered by the KDE-based inference, are constrained by high-frequency inertial measurements using on-manifold preintegration factor \cite{forster2016manifold}. For a sensor frame $S \in \{\mathcal{B}, \mathcal{L}_{1 \dots 4}\}$, assuming constant biases $\mathbf{b}_{g,k}^S$ and $\mathbf{b}_{a,k}^S$ between two consecutive frames, the rotation, velocity, and position increments are propagated as:
\begin{equation}
    \Delta \tilde{\mathbf{R}}_{k, i+1}^S = \Delta \tilde{\mathbf{R}}_{k, i}^S \text{Exp} \left( (\tilde{\boldsymbol{\omega}}_i^S - \mathbf{b}_{g,k}^S) \Delta t \right)
\end{equation}
\begin{equation}
    \Delta \tilde{\mathbf{v}}_{k, i+1}^S = \Delta \tilde{\mathbf{v}}_{k, i}^S + \Delta \tilde{\mathbf{R}}_{k, i}^S \left( \tilde{\mathbf{a}}_i^S - \mathbf{b}_{a,k}^S \right) \Delta t
\end{equation}
\begin{equation}
    \Delta \tilde{\mathbf{p}}_{k, i+1}^S = \Delta \tilde{\mathbf{p}}_{k, i}^S + \Delta \tilde{\mathbf{v}}_{k, i}^S \Delta t + \frac{1}{2} \Delta \tilde{\mathbf{R}}_{k, i}^S \left( \tilde{\mathbf{a}}_i^S - \mathbf{b}_{a,k}^S \right) \Delta t^2
\end{equation}

The residual $\mathbf{r}_{\mathcal{I}}^S$ is constructed by comparing these increments with the manifold states. The orientation, velocity, and position residuals are defined as:
\begin{equation}
    \mathbf{r}_{\Delta\mathbf{R}}^S = \text{Log} \left( (\Delta \tilde{\mathbf{R}}_{k,k+1}^S)^T {\mathbf{R}_k^S}^T \mathbf{R}_{k+1}^S \right)
\end{equation}
\begin{equation}
    \mathbf{r}_{\Delta\mathbf{v}}^S = {\mathbf{R}_k^S}^T \left( \mathbf{v}_{k+1}^S - \mathbf{v}_k^S - \mathbf{g} \Delta t_{k} \right) - \Delta \tilde{\mathbf{v}}_{k,k+1}^S
\end{equation}
\begin{equation}
    \mathbf{r}_{\Delta\mathbf{p}}^S = {\mathbf{R}_k^S}^T \left( \delta \mathbf{p} - \mathbf{v}_k^S \Delta t_{k} - \frac{1}{2}\mathbf{g} \Delta t_{k}^2 \right) - \Delta \tilde{\mathbf{p}}_{k,k+1}^S
\end{equation}
where $\delta \mathbf{p} = \mathbf{p}_{k+1}^S - \mathbf{p}_k^S$. The full IMU residual $\mathbf{r}_{\mathcal{I}}^S$ concatenates these terms with the bias random walk:
\begin{equation}
    \mathbf{r}_{\mathcal{I}}^S = \left[ {\mathbf{r}_{\Delta\mathbf{R}}^S}^T, {\mathbf{r}_{\Delta\mathbf{v}}^S}^T, {\mathbf{r}_{\Delta\mathbf{p}}^S}^T, (\Delta \mathbf{b}_{a,k}^S)^T, (\Delta \mathbf{b}_{g,k}^S)^T \right]^T
\end{equation}
$\Delta \mathbf{b}^S$ denotes the bias change between keyframes. First-order Jacobians are employed to update the preintegrated measurements during optimization without reintegration.

%%%%%%%%%%%%%%%%%%%%%%%%%%%%%%%%%%%%%%%%%%%%%%%%%%%%%%%%%%%%%%%%%%%%%%%%%%%%%%%%
\subsection{ZUPT Factor}

The ZUPT factor enforces a zero-velocity constraint on the actual contact point $f$ in the world frame $\mathcal{W}$. To bridge the kinematic gap between the Leg-IMU center $\mathcal{L}$ and the contact interface $f$, we account for the lever arm ${}^{\mathcal{L}}\mathbf{l}_{f}$ within the foot structure.

In our framework, the velocity of the contact point is modeled as a zero-velocity pseudo-measurement corrupted by Gaussian noise ${}^{\mathcal{W}}\mathbf{n}_{zv} \sim \mathcal{N}(\mathbf{0}, \boldsymbol{\Sigma}_{zv})$. The kinematic relationship is defined as:
\begin{equation}
    {}^{\mathcal{W}}\mathbf{v}_{f} = \mathbf{v}_k^{\mathcal{L}} + \mathbf{R}_k^{\mathcal{L}} \left( (\tilde{\boldsymbol{\omega}}^{\mathcal{L}}_{k} - \mathbf{b}_{g,k}^{\mathcal{L}}) \times {}^{\mathcal{L}}\mathbf{l}_{f} \right)
\end{equation}
$\tilde{\boldsymbol{\omega}}^{\mathcal{L}}_{k}$ is the raw angular velocity from the Leg-IMU, and $\mathbf{b}_{g,k}^{\mathcal{L}}$ is the corresponding gyroscope bias state at keyframe $k$.

Under the contact event $\mathcal{K}$, the zero-velocity residual $\mathbf{r}_{zv}$ is formulated as the difference between the IMU-inferred velocity and the noise-perturbed observation:
\begin{equation}
    \mathbf{r}_{zv} = {}^{\mathcal{W}}\mathbf{v}_{f} - {}^{\mathcal{W}}\mathbf{n}_{zv}
\end{equation}
By integrating this residual into the factor graph, the ZUPT constraint effectively suppresses the velocity drift of the Leg-IMUs, ensuring that the local rotation of the foot is correctly propagated to the global velocity.

%%%%%%%%%%%%%%%%%%%%%%%%%%%%%%%%%%%%%%%%%%%%%%%%%%%%%%%%%%%%%%%%%%%%%%%%%%%%%%%%
\subsection{Kinematic RelPos Factor}

The RelPos factor establishes a spatial constraint between the body center $\mathcal{B}$ and the Leg-IMU $\mathcal{L}$ utilizing the robot's forward kinematics.

Let $\tilde{\mathbf{q}}$ denote the raw joint encoder measurements. The relative position of the Leg-IMU with respect to the body, expressed locally in the body frame $\mathcal{B}$, is obtained via the forward kinematics function $f_{\mathcal{B}\mathcal{L}}(\cdot)$ as a pseudo-measurement:
\begin{equation}
    {}^{\mathcal{B}}\tilde{\mathbf{p}}_{\mathcal{L}} = f_{\mathcal{B}\mathcal{L}}(\tilde{\mathbf{q}})
\end{equation}

Concurrently, based on the manifold states being optimized, this local relative position can be analytically predicted. Given the position states of the Leg-IMU $\mathbf{p}_k^{\mathcal{L}}$ and the body $\mathbf{p}_k^{\mathcal{B}}$ in the world frame $\mathcal{W}$, alongside the body's orientation state $\mathbf{R}_k^{\mathcal{B}}$, the predicted relative position is:
\begin{equation}
    {}^{\mathcal{B}}\hat{\mathbf{p}}_{\mathcal{L}} = (\mathbf{R}_k^{\mathcal{B}})^T (\mathbf{p}_k^{\mathcal{L}} - \mathbf{p}_k^{\mathcal{B}})
\end{equation}

To account for joint sensing noise and mechanical tolerances, the kinematic observation is modeled with an additive Gaussian white noise term ${}^{\mathcal{B}}\mathbf{n}_{rel} \sim \mathcal{N}(\mathbf{0}, \boldsymbol{\Sigma}_{rel})$. The translational residual $\mathbf{r}_{rel}$ is formulated as the discrepancy between the IMU-predicted relative position and the kinematics-measured relative position:
\begin{equation}
    \mathbf{r}_{rel} = {}^{\mathcal{B}}\hat{\mathbf{p}}_{\mathcal{L}} - {}^{\mathcal{B}}\tilde{\mathbf{p}}_{\mathcal{L}} - {}^{\mathcal{B}}\mathbf{n}_{rel}
\end{equation}

By enforcing this positional residual, the RelPos factor resolves the observability ambiguity in a multi-IMU system. Specifically, when a leg enters the swing phase and loses its ZUPT constraint, its IMU biases become unobservable and prone to rapid drift. Managing thirty concurrent bias states across five IMUs would otherwise lead to severe divergence. The RelPos factor mitigates this by kinematically anchoring the airborne leg to the main body. Since the body state is continuously bounded by the active stance legs, this structural coupling implicitly transfers global observability to the swinging leg, bounding its bias random walk. Thus, compared to a standard single-body kinematic model, the RelPos factor ensures that capturing local high-frequency dynamics does not compromise overall estimation stability.

%%%%%%%%%%%%%%%%%%%%%%%%%%%%%%%%%%%%%%%%%%%%%%%%%%%%%%%%%%%%%%%%%%%%%%%%%%%%%%%%
\subsection{Maximum A Posteriori Estimation}

The proposed state estimation is fundamentally formulated as a Maximum A Posteriori (MAP) inference over a sliding window of states $\mathcal{X}$. Under the assumption of zero-mean Gaussian noise, this is equivalent to minimizing a non-linear least-squares objective defined by the sum of the squared Mahalanobis distances of all factor residuals.

As illustrated in Fig. \ref{fig:factor_topology}, to accurately reflect our graph topology, let $\mathcal{A}_k \subseteq \{\mathcal{L}_1, \dots, \mathcal{L}_4\}$ represent the dynamic subset of active stance legs registering a stable contact at keyframe $k$. The unified objective function is formulated as:
\begin{equation}
\begin{split}
    \mathcal{X}^* = \arg\min_{\mathcal{X}} \Bigg\{ &\| \mathbf{r}_{prior} \|^2_{\boldsymbol{\Sigma}_{prior}} 
    + \sum_{k \in \mathcal{K}} \sum_{S \in \{\mathcal{B}, \mathcal{L}_{1 \dots 4}\}} \| \mathbf{r}_{\mathcal{I}}^{k,S} \|^2_{\boldsymbol{\Sigma}_{\mathcal{I}}^S} \\
    &+ \sum_{k \in \mathcal{K}} \sum_{l \in \mathcal{A}_k} \left( \| \mathbf{r}_{zv}^{k,l} \|^2_{\boldsymbol{\Sigma}_{zv}} + \| \mathbf{r}_{rel}^{k,l} \|^2_{\boldsymbol{\Sigma}_{rel}} \right) \Bigg\}
\end{split}
\end{equation}
where $\mathbf{r}_{prior}$ denotes the marginalization prior. This mathematical formulation describes our system architecture: preintegration constraints are instantiated for all continuous inertial nodes to maintain global temporal continuity, while zero-velocity measurements and relative position constraints are exclusively applied to the active stance legs $l \in \mathcal{A}_k$.

The robustness of this hardware-deployed optimization relies on the physically grounded covariance matrices $\boldsymbol{\Sigma}$:
\begin{itemize}
    \item \textbf{Preintegration Covariance ($\boldsymbol{\Sigma}_{\mathcal{I}}$)} is propagated analytically from continuous-time sensor noise. Crucially, the accelerometer noise density for the Leg-IMUs is  inflated relative to the Body-IMU. This  design absorbs severe, high-frequency impact shocks during touchdown, preventing transient spikes from dominating the local inertial propagation.
    \item \textbf{ZUPT Covariance ($\boldsymbol{\Sigma}_{zv}$)} is dynamically scaled based on the continuous KDE contact probability $P_c$ instead of an empirical value. A simple exponential mapping is employed to inflate the nominal firm-contact covariance $\boldsymbol{\Sigma}_{0}$ as the contact probability decreases:
    \begin{equation}
        \boldsymbol{\Sigma}_{zv}(P_c) = \boldsymbol{\Sigma}_{0} \exp\left( \gamma (1 - P_c) \right)
    \end{equation}
    where $\gamma$ is a positive scaling hyperparameter. High confidence tightens the covariance to firmly anchor the stance foot, whereas lower probabilities proportionally inflate the uncertainty to accommodate minor slip or soft-ground deformation.
    \item \textbf{Kinematic RelPos Covariance ($\boldsymbol{\Sigma}_{rel}$)} encodes the inherent structural uncertainty in the forward kinematics map $f_{\mathcal{B}\mathcal{L}}(\cdot)$, effectively mitigating the impact of joint encoder noise and mechanical lever-arm tolerances.
\end{itemize}

This unified MAP objective is continuously optimized using an incremental fixed-lag smoother~\cite{kaess2012isam2}. Real-time onboard execution is ensured by systematically marginalizing old states outside the sliding window via the Schur complement, condensing their structural information into the prior term $\mathbf{r}_{prior}$. By weighting the residuals through these properly determined covariance profiles, the factor graph seamlessly fuses high-bandwidth inertial readings with kinematically anchored pseudo-measurements to provide consistent and robust state estimation.
%%%%%%%%%%%%%%%%%%%%%%%%%%%%%%%%%%%%%%%%%%%%%%%%%%%%%%%%%%%%%%%%%%%%%%%%%%%%%%%%

% ==================== 图 X：实验平台与环境 ====================
\begin{figure}[htbp]
    \centering
    \includegraphics[width=0.9\columnwidth]{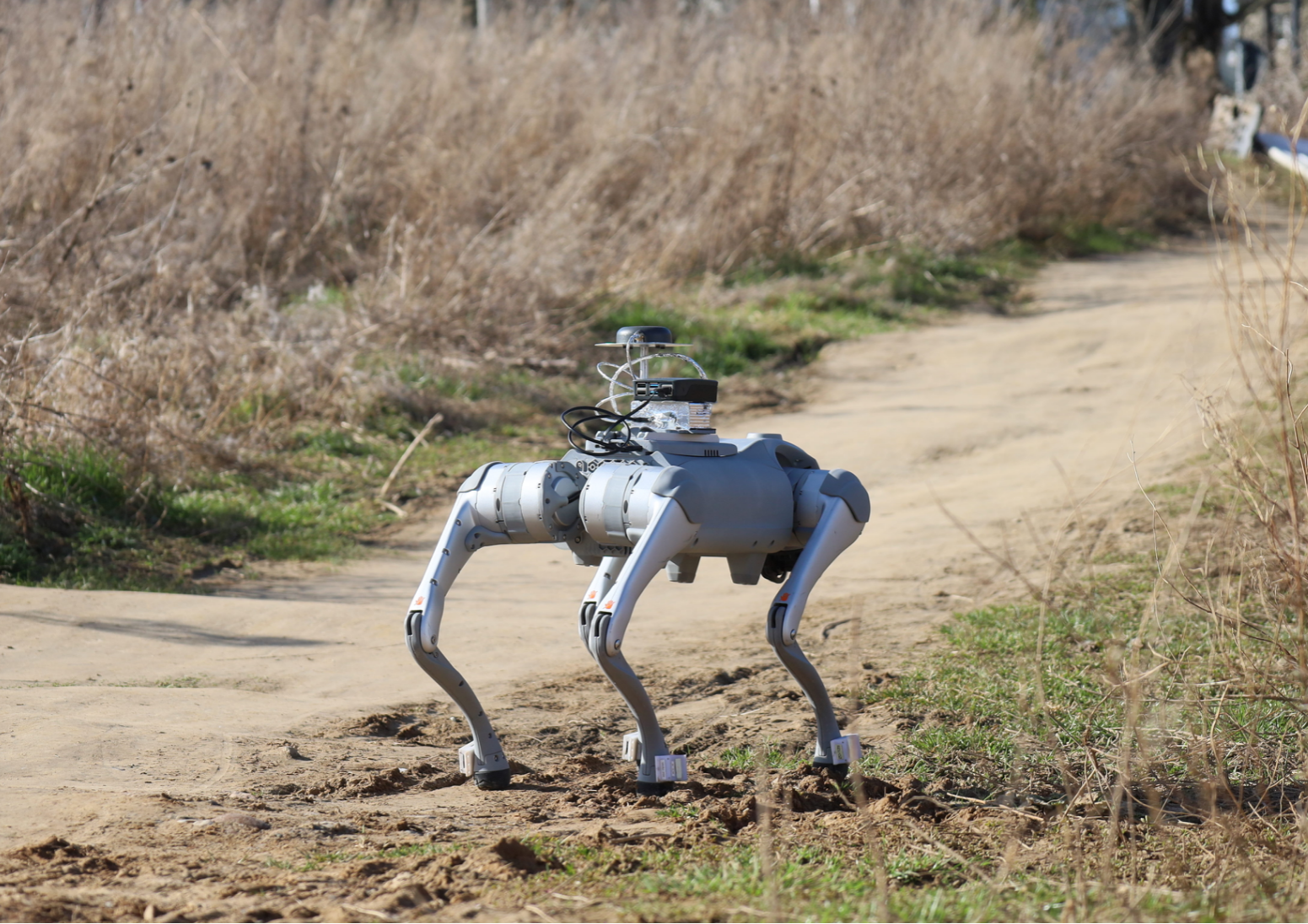}
    \vspace{-0.3cm}
    \caption{The experimental platform and hardware payload deployed in an unstructured environment (offroad). The Unitree Go2 quadruped robot is equipped with a custom multi-IMU suite, a Raspberry Pi for onboard data logging, and a GNSS antenna for ground truth acquisition.}
    \label{fig:experiment_setup}
    \vspace{-0.3cm}
\end{figure}

\section{Experiments}
This section evaluates the proposed LAG-IO framework using a publicly available real-world legged locomotion dataset. The subsequent evaluations substantiate two primary claims: first, the proposed multi-IMU graph optimization exhibits competitive state estimation accuracy against the evaluated baselines; second, the probabilistic contact detection mechanism cooperates with the graph backend to facilitate smoother trajectory transitions during unstable foot-ground interactions. Additionally, the computational efficiency of the system is analyzed to verify its real-time viability.

% ==================== 表格 1： 轨迹误差 (4种方法) ====================
\begin{table}[t]
\centering
\caption{Overall Trajectory Estimation Accuracy}
\label{tab:3d_metrics}
\begin{tabular}{llcc}
\toprule
\textbf{Sequence} & \textbf{Method} & \textbf{APE Trans. (m)} & \textbf{RPE Trans. (\%) } \\
\midrule
\multirow{4}{*}{Parking lots} 
& LegOdom & 2.95 & \textbf{1.02} \\
& Foot-INS & 4.35 & 4.32 \\
& DogLegs & 3.04 & 1.74 \\
& LAG-IO & \textbf{2.80} & 1.24 \\
\midrule
\multirow{4}{*}{Asphalt road} 
& LegOdom & 2.73 & 1.42 \\
& Foot-INS & 4.85 & 4.40 \\
& DogLegs & 3.08 & 1.94 \\
& LAG-IO & \textbf{1.93} & \textbf{1.27} \\
\midrule
\multirow{4}{*}{Construction site} 
& LegOdom & 2.17 & 0.66 \\
& Foot-INS & 1.12 & 4.04 \\
& DogLegs & 0.97 & 0.85 \\
& LAG-IO & \textbf{0.34} & \textbf{0.52} \\
\midrule
\multirow{4}{*}{Offroad} 
& LegOdom & 2.54 & 2.81 \\
& Foot-INS & 1.86 & 3.81 \\
& DogLegs & 1.71 & 1.09 \\
& LAG-IO & \textbf{1.23} & \textbf{0.63} \\
\bottomrule
\end{tabular}

\vspace{1ex} 
\parbox{\columnwidth}{\raggedright \footnotesize * APE and RPE denote the overall translational absolute pose error and relative pose error evaluated via \textit{evo} \cite{grupp2017evo}.}
\end{table}

% ==================== 表格 2： 平面轨迹误差 (2种方法) ====================
\begin{table}[t]
\centering
\caption{Planar Trajectory Estimation Accuracy}
\label{tab:2d_metrics}
\begin{tabular}{llcc}
\toprule
\textbf{Sequence} & \textbf{Method} & \textbf{APE Trans. (m)} & \textbf{RPE Trans. (\%) } \\
\midrule
\multirow{2}{*}{Parking lots} 
& DogLegs & 1.19 & 1.43 \\
& LAG-IO & \textbf{0.88} & \textbf{1.16} \\
\midrule
\multirow{2}{*}{Asphalt road} 
& DogLegs & 2.14 & 1.56 \\
& LAG-IO & \textbf{0.91} & \textbf{0.95} \\
\midrule
\multirow{2}{*}{Construction site} 
& DogLegs & 0.68 & 0.64 \\
& LAG-IO & \textbf{0.24} & \textbf{0.43} \\
\midrule
\multirow{2}{*}{Offroad} 
& DogLegs & \textbf{0.87} & 0.88 \\
& LAG-IO & 1.03 & \textbf{0.56} \\
\bottomrule
\end{tabular}

\vspace{1ex} 
\parbox{\columnwidth}{\raggedright \footnotesize * APE and RPE in this table are evaluated exclusively on the horizontal plane (XY) to eliminate the influence of unobservable vertical drift.}
    \vspace{-0.5cm}
\end{table}

% ==================== 表格 3：姿态误差 (3种方法) ====================
\begin{table}[t]
\centering
\caption{Attitude Estimation Accuracy (RMSE / MAX in Degrees)}
\label{tab:attitude_metrics}
\setlength{\tabcolsep}{1.2mm} 
\begin{tabular}{llccc}
\toprule
\textbf{Sequence} & \textbf{Method} & \textbf{Roll ($^\circ$)} & \textbf{Pitch ($^\circ$)} & \textbf{Yaw ($^\circ$)} \\
\midrule
\multirow{3}{*}{Parking lots} 
& LegOdom & 1.51 / 3.48 & 1.03 / 3.50 & 14.69 / 21.75 \\
& DogLegs & \textbf{0.62} / \textbf{1.24} & \textbf{0.42} / \textbf{1.42} & 10.03 / 16.92 \\
& LAG-IO & 0.73 / 1.85 & 0.77 / 2.88 & \textbf{7.04} / \textbf{12.48} \\
\midrule
\multirow{3}{*}{Asphalt road} 
& LegOdom & 0.69 / \textbf{1.88} & 0.83 / 2.28 & 10.05 / 16.01 \\
& DogLegs & 0.77 / 1.98 & \textbf{0.78} / \textbf{2.04} & 10.45 / 17.85 \\
& LAG-IO & \textbf{0.67} / 1.89 & 0.86 / 2.38 & \textbf{7.24} / \textbf{12.66} \\
\midrule
\multirow{3}{*}{Construction site} 
& LegOdom & 0.68 / 1.67 & 0.94 / 1.95 & 3.93 / 6.06 \\
& DogLegs & 0.68 / \textbf{1.63} & 1.07 / 1.92 & 1.40 / \textbf{3.03} \\
& LAG-IO & \textbf{0.61} / 1.69 & \textbf{0.57} / \textbf{1.76} & \textbf{0.99} / 3.81 \\
\midrule
\multirow{3}{*}{Offroad} 
& LegOdom & 0.91 / 1.90 & \textbf{0.64} / \textbf{1.73} & 29.33 / 33.17 \\
& DogLegs & 0.50 / \textbf{1.29} & 0.87 / 1.98 & \textbf{1.68} / \textbf{3.70} \\
& LAG-IO & \textbf{0.46} / 1.32 & 0.85 / 1.98 & 1.99 / 5.70 \\
\bottomrule
\end{tabular}
\vspace{-0.5cm}
\end{table}

% ==================== 图 1：2D 轨迹对比 ====================
\begin{figure}[!htbp]
    \centering
    \includegraphics[width=\columnwidth]{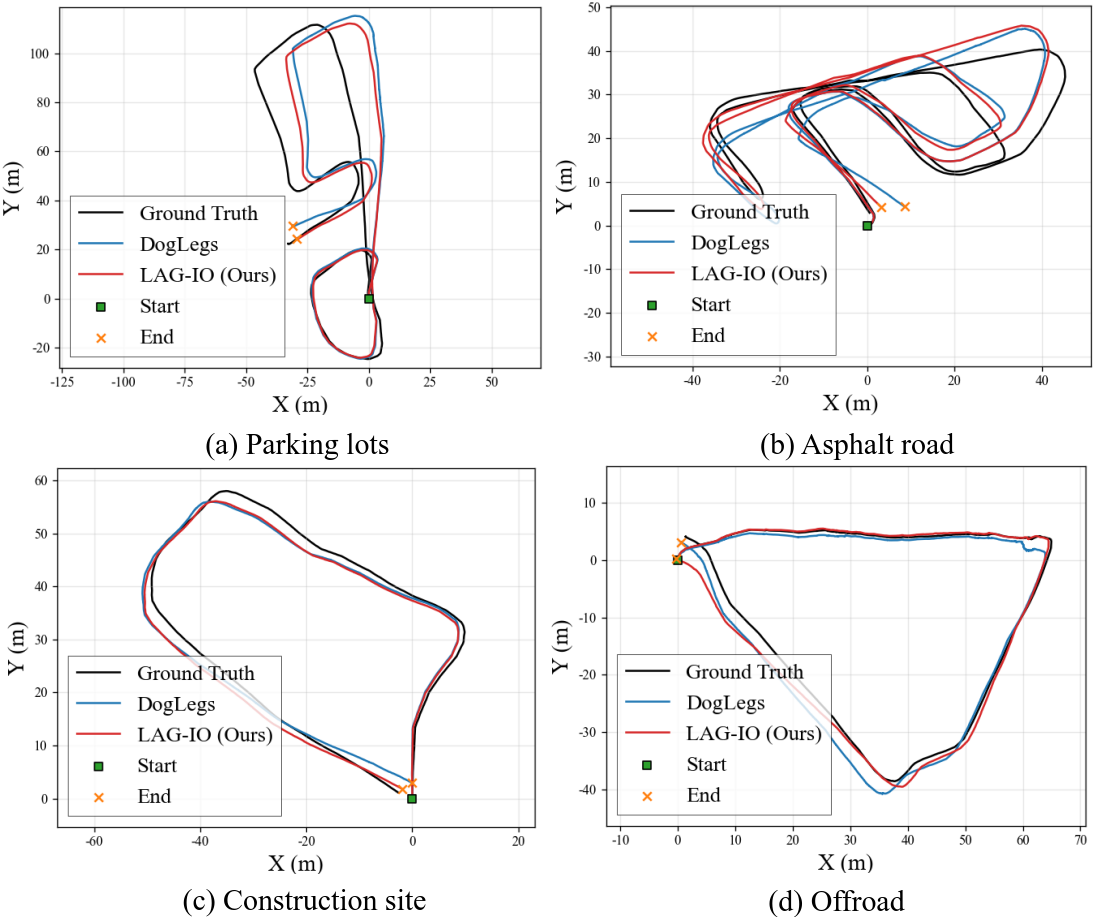}
    \vspace{-0.3cm}
    \caption{Top-down planar trajectory estimates from the DogLegs dataset across four scenarios: parking lots, asphalt road, construction site, and offroad. Overall, the proposed LAG-IO (red) maintains closer alignment with the GNSS-PPK ground truth (black) compared to the baseline (blue), which exhibits accumulated drift in certain regions.}
    \label{fig:2d_trajectories}
    \vspace{-0.3cm}
\end{figure}

% ==================== 图 2：Z轴轨迹对比 ====================
\begin{figure}[!htbp]
    \centering
    \includegraphics[width=\columnwidth]{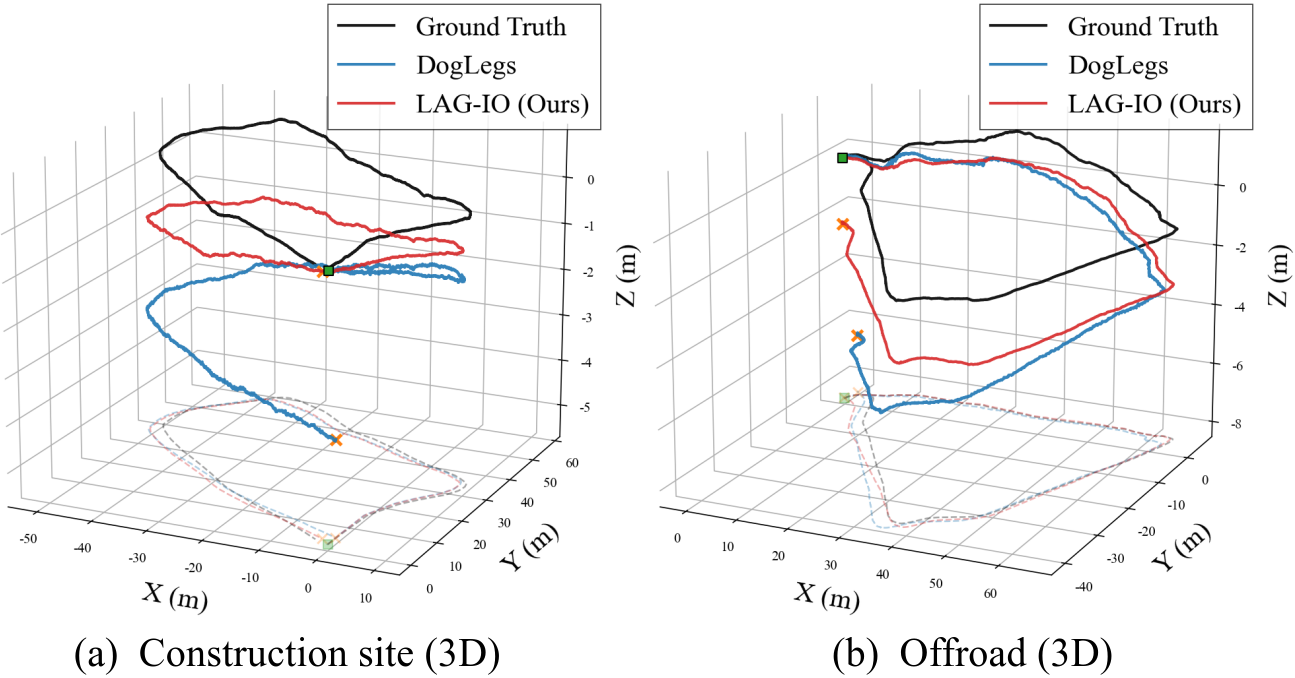}
    \vspace{-0.3cm}
    \caption{Vertical trajectory profiles for the construction site and offroad scenarios. Both methods inevitably exhibit Z-axis drift, with the proposed LAG-IO (red) maintaining a relatively milder elevation divergence compared to the DogLegs baseline (blue).}
    \label{fig:3d_z_drift}
    \vspace{-0.5cm}
\end{figure}

\subsection{Experimental Setup}
The proposed LAG-IO framework is evaluated using the dataset collected from the DogLegs \cite{wu2025doglegs}. As illustrated in Fig.~\ref{fig:experiment_setup}, the experimental platform is a Unitree Go2 quadruped robot equipped with a primary body-mounted IMU and four Leg-IMUs attached to the lower legs, close to the foot contact points. All inertial measurements rely on low-cost commercial ICM20602 sensor(3 US dollars). Both the IMU streams and joint encoder data are recorded at a nominal frequency of 200 Hz via an onboard Raspberry Pi. Spatial extrinsics among the sensor frames and the kinematic chain are calibrated prior to data collection. For quantitative evaluation, the reference ground-truth trajectories are provided by a smoothed post-processed kinematic (PPK) GNSS/IMU integrated navigation system.

The dataset includes several outdoor scenes, namely an asphalt road, parking lots, a construction site, an offroad trail, and a grassy terrain. The robot operated in a standard walking gait across all sequences, traversing typical real-world environments with natural terrain variations. Due to severe GNSS signal occlusion by surrounding trees and buildings, absolute ground truth is unavailable for the grass sequence; instead, the estimation drift is evaluated using the loop-closure error when the robot returns to its starting location.

\subsection{State Estimation Performance}

We evaluate the state estimation performance of the proposed LAG-IO framework against three existing proprioceptive estimators:
\begin{itemize}
    \item \textbf{LegOdom.} The traditional EKF-based leg odometry using a body-mounted IMU and joint encoders \cite{bloesch2013state}.
    \item \textbf{Foot-INS.} A single foot-mounted IMU-based localization system \cite{niu2022footins}.
    \item \textbf{DogLegs.} A tightly-coupled multi-IMU EKF baseline that fuses a body IMU, leg IMUs, and forward kinematics \cite{wu2025doglegs}.
\end{itemize}
Since the optimization and marginalization mechanisms yield a sparser state trajectory for the graph, both the ground truth and the baseline trajectories are downsampled to align with the graph's timestamps prior to evaluation.

Fig.~\ref{fig:2d_trajectories} and Fig.~\ref{fig:3d_z_drift} present the top-down planar trajectories and vertical profiles, respectively. To maintain visual clarity, we illustrate the proposed LAG-IO alongside the most competitive multi-IMU baseline, DogLegs. Certain sequences exhibit initial heading deviations caused by slightly inaccurate attitude and IMU bias calibrations during the stationary initialization. Furthermore, due to the lack of absolute elevation referencing, all methods experience Z-axis drift. Despite this, LAG-IO demonstrates relatively milder elevation divergence and maintains higher geometric fidelity to the ground truth. This mitigated vertical drift is primarily attributed to the RelPos factors establishing a tightly constrained kinematic chain.

For quantitative evaluation, Tab.~\ref{tab:3d_metrics} and Tab.~\ref{tab:2d_metrics} report the overall and planar translational absolute and relative pose errors (APE and RPE) evaluated via \textit{evo} \cite{grupp2017evo}. As unobservable vertical drift inflates the overall spatial errors, the isolated planar metrics in Tab.~\ref{tab:2d_metrics} provide a clearer assessment of odometry performance. By iteratively re-optimizing the relative position constraints within the sliding window, our framework effectively suppresses the multi-directional drift of foot IMUs, resulting in superior tracking accuracy.

Moving to orientation performance, Tab.~\ref{tab:attitude_metrics} details the attitude estimation accuracy. All evaluated frameworks demonstrate comparable, sub-degree accuracy in observable roll and pitch states, which are globally constrained by gravity. Regarding the unobservable heading state, LAG-IO maintains highly competitive orientation accuracy alongside the baselines.

\begin{figure}[t]
    \centering
    \includegraphics[width=0.8\columnwidth]{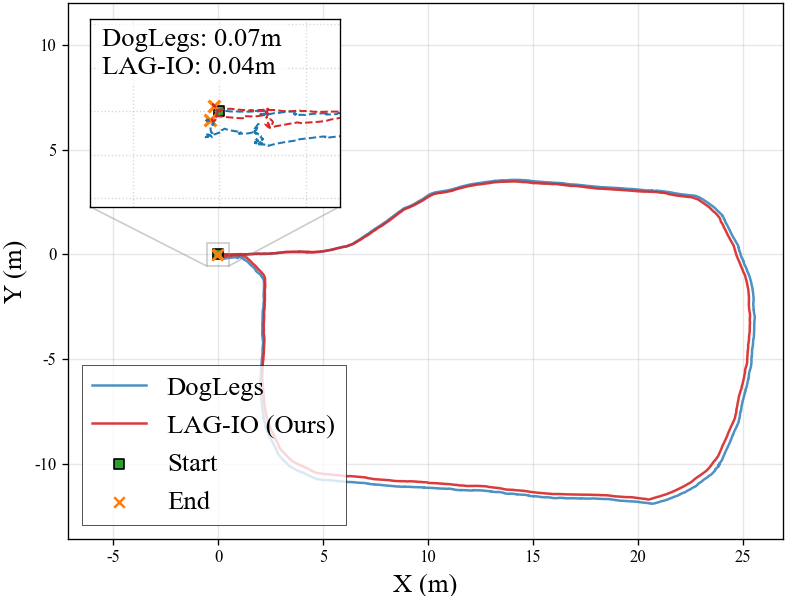}
    \caption{Trajectory evaluation in the GNSS-denied grass scenario. The proposed LAG-IO achieves a final loop-closure error of 0.04 m, outperforming the DogLegs baseline.}
    \label{fig:loop_closure}
    \vspace{-0.5cm}
\end{figure}

\begin{figure*}[!htbp]
    \centering
    \includegraphics[width=0.9\textwidth]{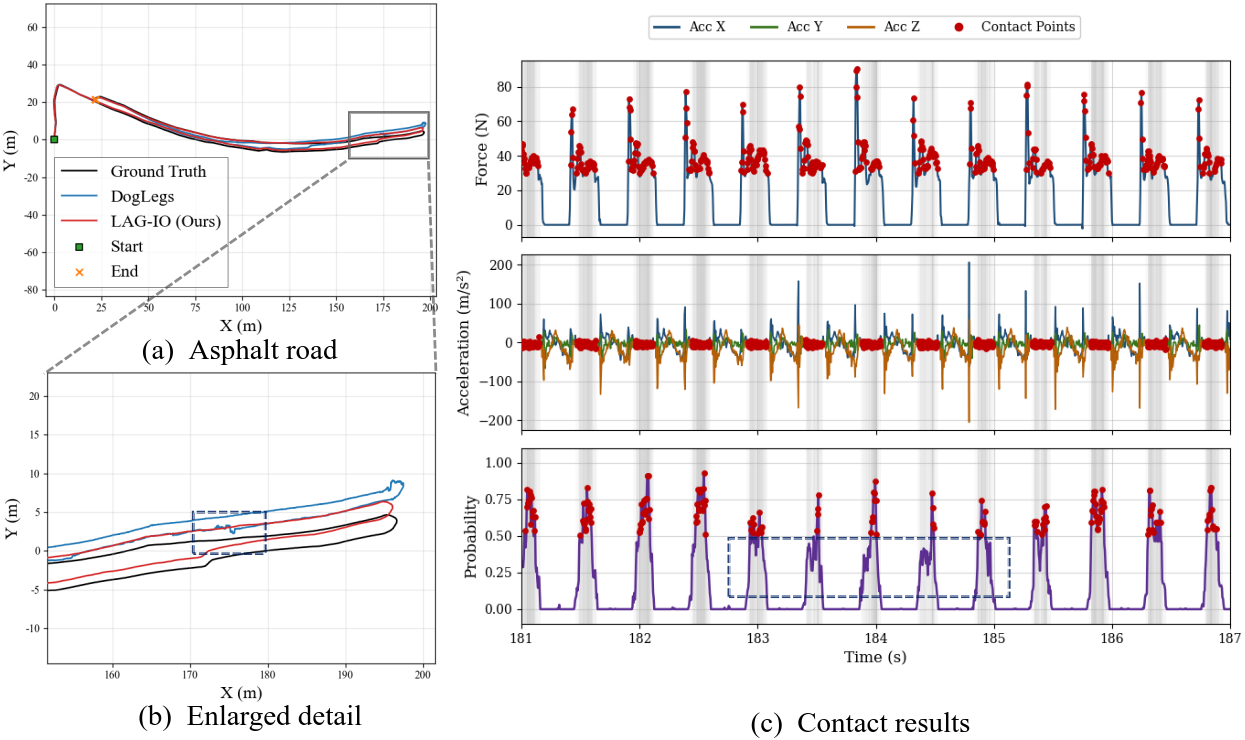}
    \vspace{-0.3cm}
    \caption{Visualization of contact detection results during a trajectory jitter event on the asphalt road sequence. (a) Full 2D trajectory. (b) Enlarged detail highlighting localized estimation jitters and fluctuations in the baseline. (c) Temporal contact detection subplots: foot force (top), baseline likelihood ratio test (middle), and KDE-based probability (bottom). The red dots indicate contact triggers thresholded by these three mechanisms. The dashed box in (c) illustrates the specific contact probability conditions corresponding to the trajectory jitter boxed in (b), revealing a drop in contact stability.}
    \label{fig:contact_analysis}
    \vspace{-0.5cm}
\end{figure*}

Finally, Fig. \ref{fig:loop_closure} illustrates the estimation performance on a grassy terrain lacking GNSS references. In this scenario, the robot navigates a long loop over a slip-prone surface. The accumulated drift is quantified by the loop-closure distance upon returning to the start. DogLegs finishes with a terminal error of $0.07$\,m, whereas LAG-IO achieves a residual error of only $0.04$\,m, validating its dead-reckoning reliability in perceptually degraded environments.

\subsection{Impact of Probabilistic Contact Detection}

We evaluate the impact of the probabilistic contact formulation using the asphalt road sequence. As illustrated in Fig. \ref{fig:contact_analysis}a, this scenario features a long and narrow path terminating in sharp maneuvers, where unstable foot-ground slipping is highly likely to occur while turning. 

As highlighted in the enlarged detail of Fig. \ref{fig:contact_analysis}b, the DogLegs baseline exhibits localized trajectory jitters and geometric fluctuations during the sharp turn. In contrast, LAG-IO effectively mitigates these distortions, tracking the ground truth smoothly. The temporal subplots (Fig. \ref{fig:contact_analysis}c) reveal the adaptive mechanism driving this improvement. During high-frequency impacts or micro-slips, traditional force sensors may still report high-magnitude readings, erroneously triggering rigid ZUPT constraints that inject bad information. At the exact timestamps where the baseline jitters (dashed boxes), force data indicates active contact, yet the KDE-based probability noticeably drops. This transient loss of stability is precisely what binary force detection fails to capture.

Within the LAG-IO framework, this lower probability is utilized to scale the covariance of the ZUPT factors to decrease their confidence. Consequently, the factor graph relaxes the stationarity constraint and allows the optimization to rely more on inertial measurements and relative kinematics to bridge the unstable transition. Since the robot maintained a consistent walking gait throughout the dataset, the bandwidth parameter $d_j$ was kept constant. Tuning $d_j$ for vastly different vibration profiles in dynamic gaits (e.g., trotting), alongside quantitatively analyzing contact stability using high-frequency motion capture references, remains a primary focus for our future research.

\subsection{Computational Efficiency}
This section analyzes the computational efficiency of the proposed LAG-IO. All experiments were conducted on a computational platform equipped with an Intel i7-14650HX CPU and 32 GB of RAM, with the processor configured to a clock frequency of 2.20 GHz. To evaluate the real-time performance, all algorithmic modules were executed in a persistent process using a single computation thread.

The \textit{Factor Creation} stage, responsible for constructing multi-IMU preintegration, KDE-scaled ZUPT, and kinematic RelPos constraints, requires an average computation time of $0.15$\,ms per keyframe. The \textit{Optimization \& Marginalization} process is performed via the iSAM2 incremental solver within a fixed-lag sliding window. Specifically, the smoother marginalizes old states based on their timestamps with a lag threshold of $0.2$\,s, maintaining an active optimization window of approximately $10$ keyframes. This combined optimization step takes $4.67$\,ms on average. Since the graph optimization is not triggered for every incoming high-frequency IMU frame, the overall framework achieves an average processing time of $1.67$\,ms per IMU measurement. These results demonstrate that the proposed system is capable of reliably operating at a processing frequency exceeding $200$\,Hz, fully satisfying the real-time requirements for proprioceptive state estimation on legged robots.
%%%%%%%%%%%%%%%%%%%%%%%%%%%%%%%%%%%%%%%%%%%%%%%%%%%%%%%%%%%%%%%%%%%%%%%%%%%%%%%%
\section{Conclusion}

This paper presented LAG-IO, a tightly-coupled factor graph optimization framework for multi-IMU proprioceptive state estimation in legged robots. By migrating the sensor fusion architecture from traditional recursive filtering to a sliding-window graph backend, our approach addresses the limitations associated with fixed linearization points. To handle unstable foot-ground interactions, we integrated a continuous probabilistic contact detection module based on kernel density estimation. This mechanism adaptively scales the zero-velocity constraints, relaxing them during micro-slips to prevent the injection of erroneous kinematic updates. Furthermore, kinematic relative position factors were formulated to structurally anchor the independent leg inertial states to the robot's main body. 

Experiments demonstrated that, compared to the evaluated baselines, LAG-IO achieves higher planar trajectory accuracy, reduces unobservable vertical divergence, and provides smoother trajectory transitions during unstable contacts, while maintaining competitive attitude observability. Future research will focus on a more exhaustive exploration of the adaptive probability mapping utilizing high-frequency motion capture references, alongside the integration of exteroceptive perception (e.g., vision or LiDAR) to achieve resilient autonomous navigation in unstructured environments.
%%%%%%%%%%%%%%%%%%%%%%%%%%%%%%%%%%%%%%%%%%%%%%%%%%%%%%%%%%%%%%%%%%%%%%%%%%%%%%%%
\section*{ACKNOWLEDGMENT}
We thank Dr. Yibin Wu for his assistance in running DogLegs and helpful discussion on the experiments.
%%%%%%%%%%%%%%%%%%%%%%%%%%%%%%%%%%%%%%%%%%%%%%%%%%%%%%%%%%%%%%%%%%%%%%%%%%%%%%%%

\bibliographystyle{ieeetr}
\bibliography{references}

\end{document}